\documentclass{article}

\usepackage{microtype}
\usepackage{graphicx}
\usepackage{subfigure}
\usepackage{booktabs} 
\usepackage{multirow} 
\usepackage{hyperref}
\usepackage{mathrsfs}
\usepackage[most]{tcolorbox}
\definecolor{keywordcolor}{rgb}{0.7, 0.1, 0.1}   
\definecolor{tacticcolor}{rgb}{0.0, 0.1, 0.6}    
\definecolor{commentcolor}{rgb}{0.4, 0.4, 0.4}   
\definecolor{symbolcolor}{rgb}{0.0, 0.1, 0.6}    
\definecolor{sortcolor}{rgb}{0.1, 0.5, 0.1}      
\definecolor{attributecolor}{rgb}{0.7, 0.1, 0.1} 
\def\lstlanguagefiles{lstlean.tex}

\usepackage[accepted]{icml2025}

\usepackage{amsmath}
\usepackage{amssymb}
\usepackage{mathtools}
\usepackage{amsthm}

\usepackage[capitalize,noabbrev]{cleveref}

\theoremstyle{plain}

\theoremstyle{definition}

\theoremstyle{remark}

\usepackage{subcaption}
\usepackage[textsize=tiny]{todonotes}
\def\lstlanguagefiles{lstlean.tex}

\newcommand{\thiswork}{InternLM2.5-StepProver}

\icmltitlerunning{InternLM2.5-StepProver: Advancing Automated Theorem Proving via Critic-Guided Search}

\begin{document}

\twocolumn[
\icmltitle{InternLM2.5-StepProver: Advancing Automated Theorem Proving via Critic-Guided Search}

\icmlsetsymbol{equal}{*}

\begin{icmlauthorlist}
\icmlauthor{Zijian Wu}{equal,shai,cuhk}
\icmlauthor{Suozhi Huang}{equal,tsinghua}
\icmlauthor{Zhejian Zhou}{shai,usc}
\icmlauthor{Huaiyuan Ying}{shai,tsinghua}
\icmlauthor{Zheng Yuan}{cuhk}
\icmlauthor{Wenwei Zhang}{shai}
\icmlauthor{Dahua Lin}{shai,cuhk}
\icmlauthor{Kai Chen}{shai}
\end{icmlauthorlist}

\icmlaffiliation{shai}{Shanghai AI Laboratory, Shanghai, China}
\icmlaffiliation{cuhk}{The Chinese University of Hong Kong, Hong Kong, China}
\icmlaffiliation{tsinghua}{Tsinghua University, Beijing, China}
\icmlaffiliation{usc}{University of Southern California, Los Angeles, USA}

\icmlcorrespondingauthor{Zijian Wu}{wuzijian@pjlab.org.cn}
\icmlcorrespondingauthor{Kai Chen}{chenkai@pjlab.org.cn}

\icmlkeywords{Machine Learning, Automated Theorem Proving, LEAN, Critic-Guided Search}

\vskip 0.3in
]



\printAffiliationsAndNotice{\icmlEqualContribution} 

\begin{abstract}
Large Language Models (LLMs) have emerged as powerful tools in mathematical theorem proving, particularly when utilizing formal languages such as LEAN.
A prevalent proof method involves the LLM prover iteratively constructing the proof tactic by tactic, typically following a best-first search scheme.
However, this method often ignores the critical preference information inside the existing tactic trajectories, hindering the search for deeper proofs.
We propose an intuitive yet effective method, which utilizes a critic model to capture the preference information and to guide the search of the prover model at runtime.
Given the prover-critic framework, a large-scale expert iteration with more than 20,000 CPU days is then applied to further fine-tune the prover and the critic.
The trained {\thiswork} critic significantly boosts the performance of the prover model ($59.4\%  \to 65.9\%$).
We also analyze the impact of the critic on various aspects of the theorem proving process during expert iteration, providing insights into its effectiveness.
The models and the discovered proofs will be open-sourced.
\end{abstract}

\section{Introduction}

Automated theorem proving has been a challenging topic in artificial intelligence \citep{pfenning2004automated, zheng2021minif2f, wu2022autoformalization, polu2022formal} which requires complex reasoning and a deep understanding of mathematics.
AlphaProof\footnote{\url{https://deepmind.google/discover/blog/ai-solves-imo-problems-at-silver-medal-level/}} has demonstrated remarkable progress by achieving silver-medal performance on International Mathematical Olympiad problems using the LEAN 4 proof assistant, particularly excelling in number theory and algebra. The training regime of AlphaProof, based on the AlphaZero methodology \citep{silver2017mastering}, encompasses 100 million formal mathematics problems—a scale that significantly surpasses previous efforts \citep{polu2022formal, lample2022hypertree, xin2024deepseekproveradvancingtheoremproving, xin2024deepseekproverv15harnessingproofassistant}. 

Existing open-source methods typically train a language model on tuples of (proof state, next tactic), followed by best-first tree search to find proofs \citep{polu2020generative, polu2022formal, lample2022hypertree, yang2024leandojo, lin2024leanstarlearninginterleavethinking,wu2024leangithubcompilinggithublean}. The term "best-first" refers to relying on internal indicators from the prover model (e.g., the log-probability scores of generated tactics) to guide the search. An observation is that such indicators become unreliable when searching for deeper proofs, leading to performance degradation. Experimental results from previous studies \citep{wu2024leangithubcompilinggithublean} have shown that even when the prover model is trained on extensive formal language corpora, the ability to discover deeper proofs remains difficult to acquire. For example, although InternLM2-StepProver achieves 48.8\% on miniF2F-test, the longest proof it discovers contains only 8 tactics—a stark contrast to typical tree searches that require 100 to 1000 steps.

Inspired by recent advances in informal mathematical reasoning \citep{lightman2023letsverifystepstep,xu2024chatglmmathimprovingmathproblemsolving}, our key insight is that the tree structure constructed during the search process is naturally suited for process supervision. The search trajectories inherently provide the data needed to train the supervision model—which we term the \emph{critic} model. Therefore, a prover-critic framework is built in which the critic model estimates the "value" of each state and decides which states to explore further.

Building on this framework, our work then leverages the Lean-workbook \citep{ying2024leanworkbooklargescalelean}, the largest open-source problem collection available, to conduct systematic expert iteration and analyze proving strategies at scale. We initiate \thiswork-critic by collecting roughly 8000 preference pairs from the search trajectories when InternLM2-StepProver is evaluated on its training sets. We then perform expert iteration on Lean-workbook. Our extensive experimentation, consuming over 20,000 CPU days, yielded several key contributions to the challenge of automated theorem proving:

\begin{itemize}
    \item We present a novel approach that significantly enhances theorem proving capabilities by utilizing a separate critic model to guide the search for deeper proofs. Experiments demonstrate that the critic model boosts the performance of the prover model from $59.4\%$ to $65.6\%$ on miniF2F (when used without the vanilla best-first search method) or to $65.9\%$ when combined with it.
    \item We provide an example of massive expert iteration, demonstrating the effectiveness of expert iteration on synthetic datasets. The enlarged Lean-workbook solution set will be open-sourced.
    \item We identify a log-linear relationship between the number of proved problems and both proof length and computational resources, providing valuable insights for resource allocation in future work.
\end{itemize}

\begin{figure*}[t!] 
    \centering
    \begin{minipage}[t]{0.43\textwidth}
        \centering
        \includegraphics[width=\textwidth]{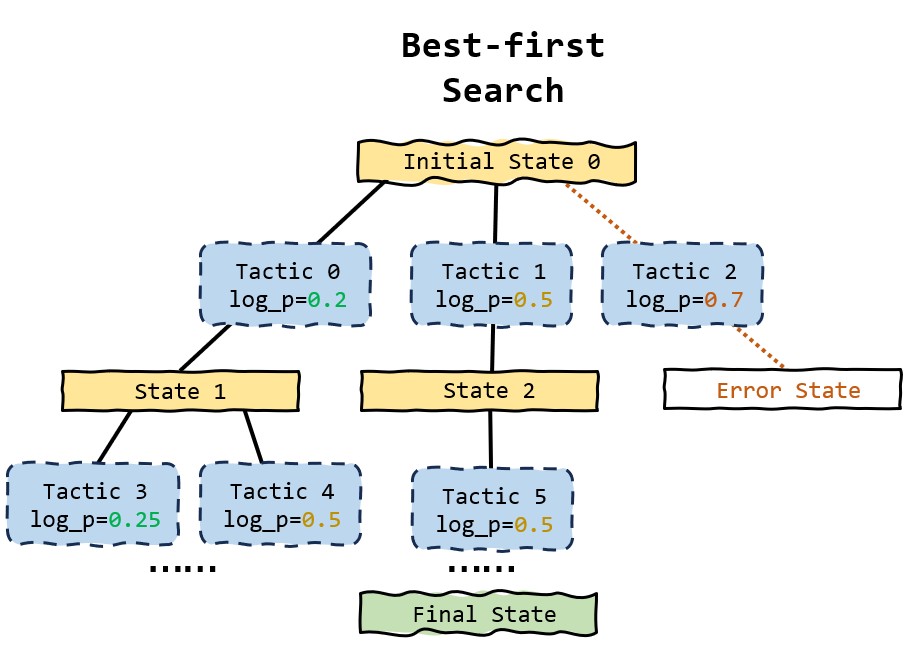}
        \label{fig:bf_search}
    \end{minipage}
    \hfill 
    \begin{minipage}[t]{0.53\textwidth}
        \centering
        \includegraphics[width=\textwidth]{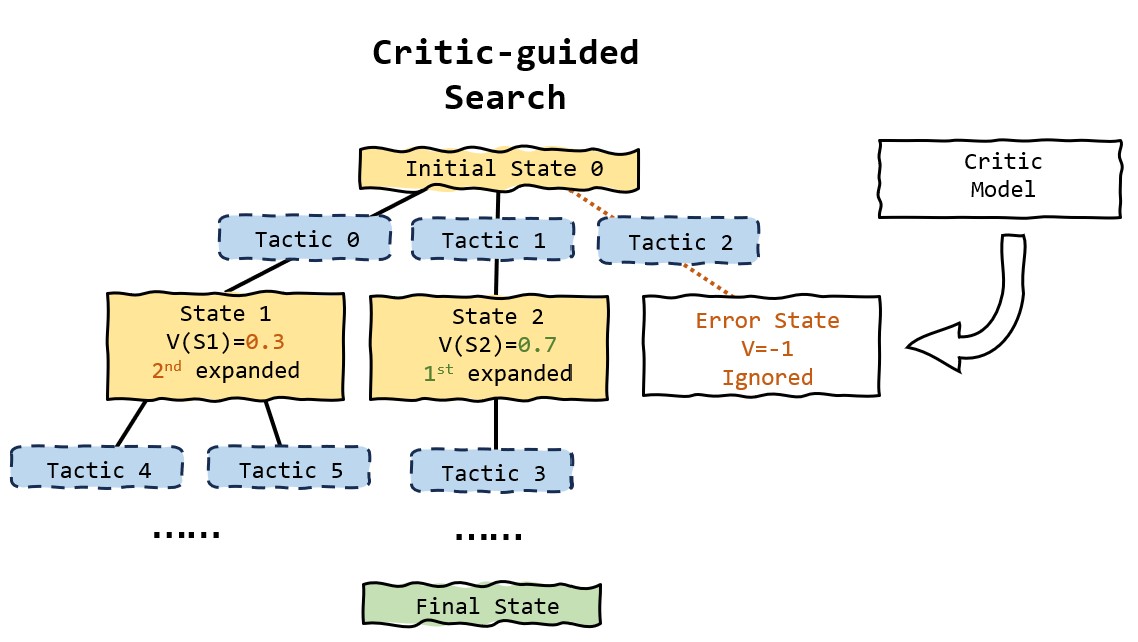}
        \label{fig:cg_search}
    \end{minipage}
    \caption{Comparison of Best-first Search and Critic-Guided Search methods. Left: Best-first search relies solely on log probability scores from tactic logits, under-utilizing intermediate state information. Right: Critic-Guided search employs a critic model to re-rank states, enabling more informed state expansion decisions. Note that a lower log probability score indicates a higher priority, whereas a higher critique score also signifies increased priority.
}
    \label{fig:search_comparison}
\end{figure*}

\section{Methods}

We introduce \textbf{critic-guided search}, a novel framework that enhances automated theorem proving by incorporating critical evaluation mechanisms into the proof search process, to address key limitations of existing best-first search methods through structured critique generation and expert iteration \citep{expertit,polu2022formal}.
The methodology consists of two main components: First, we present a critic-guided search algorithm and compare it with traditional best-first search approaches (\S\ref{ssec:prelim}). Second, we describe our expert iteration framework that iteratively refines the proof search strategy through bootstrapping processes.

\subsection{Search Methods}\label{ssec:prelim}
\paragraph{Best-First Search.} The most widely adopted approach for evaluating the theorem proving capabilities of language models $M$ involves employing best-first search algorithms, as demonstrated in GPT-f \citep{polu2020generative,azerbayev2023llemma}. This method maintains a collection of all unexpanded states $s_i$. At each iteration, the algorithm selects the optimal state $s_i$ for expansion and employs the language model to generate $S$ candidate tactics $t_{i, 1 \cdots S}$ for the current state $s_i$. For each valid tactic $t_{i,j}$, a new state is derived by executing tactic $t_{i,j}$ on state $s_i$.

Following established conventions~\citep{polu2020generative,yang2024leandojo}, the state with the highest negative log-probability is considered optimal. Specifically, we select the state $s^*$ that satisfies:

$$s^* = \arg\max_{s_i \in \mathcal{S}} \left\{ \sum_{j=0}^{i-1} -\log p(t_j | s_j)   \right\}$$

where $\mathcal{S}$ represents the set of all unexpanded states, $(s_0, t_0), \cdots, (s_{i-1}, t_{i-1})$ denotes the proof trajectory leading to state $s_i$, and $\log p(t_j | s_j)$ represents the average log probability of each generated token conditioned on the state. The algorithm expands up to $N$ states, achieving successful proof search upon reaching any proof state with no remaining goals. Multiple search attempts ($K$ times) can be performed to calculate a pass rate $pass@K$.  However, analysis of the search trajectories of our models indicates that the best-first search method exhibits poor performance (as shown in Fig.~\ref{fig:len-distribution}). Thus, we hypothesize that average log probabilities may not serve as an effective metric to guide the heuristic search process.

\begin{figure}[h]
    \centering
    \begin{subfigure}
        \centering
        \includegraphics[width=\linewidth]{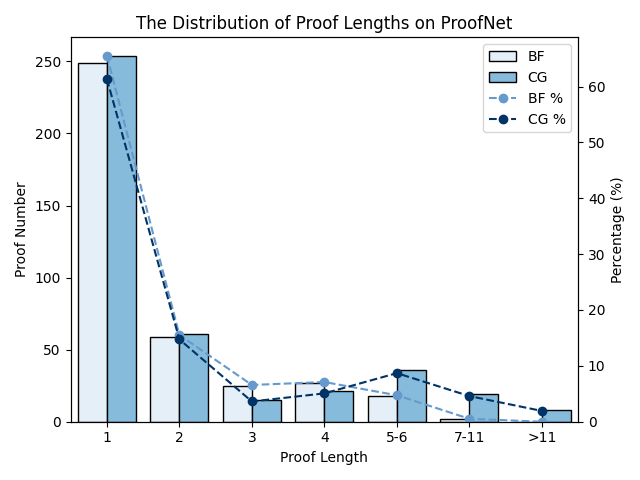}
    \end{subfigure}
    \hspace{0.05\linewidth}%
    \begin{subfigure}
        \centering
        \includegraphics[width=\linewidth]{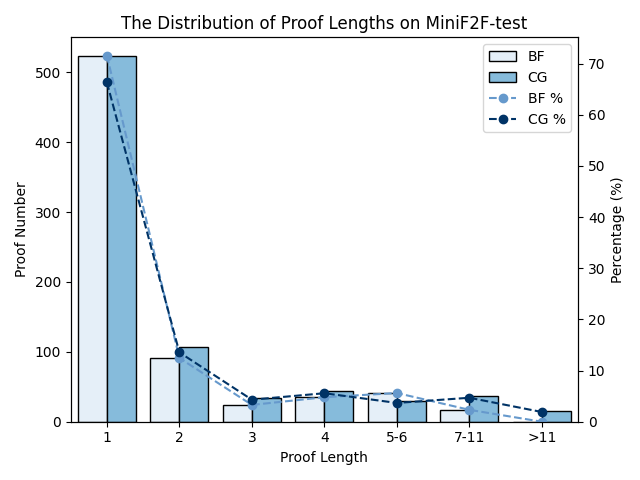}
    \end{subfigure}
    \caption{Critic-guided search finds more deep proofs compared to best-first-search in ProofNet and miniF2F-test. We calculate the proof length based on the number of tactics. Only the shortest 5 proofs are considered for each solved problem.}
    \label{fig:proof_depth}
\end{figure}

\paragraph{Critic-Guided Search}
\label{sec:critic}
Recent literature~\citep{lin2024leanstarlearninginterleavethinking} indicates that best-first search (BFS) methods may be suboptimal for formal proving. A primary advantage of tree search strategies, when compared to whole-proof generation, is their capacity to leverage critical information from intermediate prover states. However, best-first search algorithms, as they only look at the average log probability of \textit{tactics}, often fail to fully capitalize on this advantage, as shown in Fig~\ref{fig:search_comparison}.
Therefore, using best-first-search with log-probability scores seldom leads to deep proofs and limits the proving ability of our model, which is consistent with our experiments (shown in Fig~\ref{fig:proof_depth}).
Therefore, we choose to train a critic model \cite{lample2022hypertree, polu2022formal} to better guide our prover model for proof generation.

The critic model ($V$) uses the proof state ($s$) as the input and outputs a scalar ($V(s) \in \mathbb{R}$). At each iteration, the algorithm selects the optimal state $s_i$ for expansion by querying the critic model. Then it employs the prover model to generate candidate tactics.
We train our critic model in a preference style which is similar to reward model training in RLHF \citep{rlhf} instead of binary targets (the state can be proved or not). Two types of preference pairs are created in the training process:

\begin{itemize}
\item \textbf{Path Pairs}: For a successful proof path from the root (initial proof state) to \texttt{no\_goals}, we hypothesize that a state close to the target is always better. Thus we create preference pairs where the positive example is a child state closer to \texttt{no\_goals} and the negative example is the parent state closer to the initial root state (i.e. $V(s_t) < V(s_{t+\Delta})$), which implicitly assumes that any legal tactic on a successful path toward \texttt{no\_goals} yields a positive reward $r$.
For a path of length n, this methodology allows us to generate at most $n\choose 2$ pairs of positive and negative examples.

\item \textbf{Sibling Pairs}: We construct preference pairs consisting of a state on the successful path (positive example) and its sibling state (negative example) (i.e. $V(s_{sibling}) < V(s_t)$). Sibling states are defined as child nodes of the same parent that did not lead to \texttt{no\_goals}. This design is based on the principle that the state on the successful path ($s_t$) is preferable to its sibling state ($s_{sibling}$), which did not lead to a proof.

\end{itemize}
\subsection{Expert Iteration}
\paragraph{Bootstrapping}
The proposed approach utilizes the InternLM2-StepProver\citep{wu2024leangithubcompilinggithublean}, our latest model for formal reasoning, as the foundational prover model, along with an initial dataset. This dataset is aggregated from four distinct sources: the miniF2F-train split, the mathlib dataset, the Lean-Workbook, and the Lean-Github dataset. Collectively, these sources furnish the initial states $s^i$, each representing a theorem to be proven.
For the initialization of the critic model, the prover model is initially tasked with generating $K$ solution samples (trajectories) per theorem. Following this, a preference dataset is curated by sampling preference pairs from successful search trajectories; this dataset is then used for training the prover model. The overall iterative procedure is executed multiple times. The process terminates when the critic-guided search method achieves performance superior to that of standard (vanilla) methods on the miniF2F-train benchmark, which serves as an indicator that the critic model is stabilized.

\paragraph{Expert Iteration on Lean-Workbook}

With our initialized prover and critic models, we perform expert iteration on the Lean-Workbook dataset. Due to limitations in auto-formalization accuracy, it is recognized that some formalized propositions might be incorrect. Therefore, every formalized proposition includes its negated version in the dataset, following the paradigm outlined in \cite{xin2024deepseekproveradvancingtheoremproving}. Proofs are searched using best-first and critic-guided algorithms by generating a tactic as an action \citep{polu2022formal,azerbayev2023llemma,wu2024leangithubcompilinggithublean}.

Initially, we conduct a rapid scan of the entire Lean-workbook-plus dataset using a relatively small search budget (i.e., a maximum of 10 iterations per problem and a time limit of 50 seconds). The discovered proofs are added to the training set, and the solved problems, along with their negated statements, are removed from the dataset. This process helps us identify statements that are inherently unprovable, thereby enhancing the efficiency of the iteration.

This process is then repeated over multiple rounds, gradually increasing the search budget for subsequent evaluations until a predefined upper bound (at most 2000 iterations and 3600 seconds per problem) is reached. After each round, we retrain our prover and critic models using an expanded set of successful proof trajectories. Since some found proofs are ill-formed and contain many irrelevant proof steps with higher CPU consumption, we continue to search for proofs for these problems, aiming to use shorter and more direct proofs to improve our models.

After model iterations, we use the critic model to re-estimate all unproven statements. Next iteration will then focus on the top 50\% of problems that the critic model indicates are most likely to be solvable.

\paragraph{Prover Model Objective}

The traditional \textit{proofstep} objective, used by \textit{GPT-f} \citep{polu2020generative}, generates a \texttt{PROOFSTEP} (a Lean tactic) given a \texttt{GOAL} (current Lean tactic state) and the current \texttt{DECLARATION} (the Lean theorem name to be proved). The actual prompt used by GPT-f includes an additional declaration field, i.e., \texttt{DECL <DECLARATION> \textbackslash{n}GOAL <GOAL> \textbackslash{n}PROOFSTEP <PROOFSTEP>}. 
However, such prompts, though easy to integrate with existing deployment frameworks, lack information regarding the previous proof contents. Hoping to improve the reasoning performance in deep search trees, we augment our prompt template with ongoing proof context. The format of the prompt is modified to include the previous tactics leading to the state in a field called \texttt{PROOF\_BEFORE}. An example of the prompt template is shown in Fig.\ref{tab:dataexample}.

\begin{figure}
    \centering
\begin{lstlisting}[language=LEAN]
---
NAME: square_sub_one_divisible_eight
---
PROOF_BEFORE: rw [h, pow_two]
---
STATE_BEFORE: m n : ℕ
h : n = 2 * m + 1
⊢ 8 | (2 * m + 1) * (2 * m + 1) - 1
---
TACTIC:
\end{lstlisting}
\begin{lstlisting}[language=LEAN]
rw [← Nat.mod_add_div (2 * m + 1) 8]
\end{lstlisting}
    \caption{An example of the prompt used by the prover model.}
    \label{tab:dataexample}
\end{figure}

\section{Experiments and Results}

We evaluate our approach on Lean-workbook-plus~\cite{ying2024leanworkbooklargescalelean}, one of the largest auto-formalized problem sets in Lean 4, which provides comprehensive coverage of diverse mathematical reasoning tasks. The integration of critic mechanisms enables our system to discover deeper proofs while maintaining computational efficiency. We trained \thiswork~and \thiswork-critic on Lean-Workbook with more than 20000 CPU hours of expert iteration. The models are then evaluated on miniF2F\citep{zheng2021minif2f} and ProofNet\citep{azerbayev2023proofnetautoformalizingformallyproving}.

Experimental results show that both critic-guided search and expert iteration boost the theorem proving capabilities of the prover model. The critic-guided search techniques significantly enhance the prover model in its ability to find deeper proofs.
We describe our experiments and findings in detail below.

\subsection{Benchmark Performance of \thiswork}
A comprehensive analysis of \thiswork\ is conducted on several standard formal benchmarks, in comparison with our previous model InternLM2-StepProver, as well as a number of other frontier language models, to exhibit the strength of our approach.

\begin{table*}[!t]
\begin{center}
\caption{
Compared with other baselines on the miniF2F \citep{zheng2021minif2f} dataset. BF represents best-first-search and CG represents critic-guided search.
}
\label{tab:main_results} 
\small
\begin{tabular}{lcccc}
\toprule
    Method & Model size & Pass & miniF2F-valid & miniF2F-test \\
     \midrule
    \multicolumn{5}{l}{\textit{Whole-Proof Generation Methods}} \\
    \midrule
    TheoremLlama \citep{wang2024theoremllamatransforminggeneralpurposellms} & - & cumulative & $36.5\%$ & $33.6\%$ \\
 {DeepSeek-Prover \citep{xin2024deepseekproveradvancingtheoremproving}} & {7B}& $128$ & - & $46.1\%$ \\
     &  & $16\times 4096$ &- & $50.0\%$ \\
    {DeepSeek-Prover-V1.5-RL} & {7B} 
     & 32 & - & $50.0\%$ \\
     & & 64               &-& $50.7\%$ \\
     & & $128$            &-& $51.6\%$ \\
     & & $3200$           &-& $54.9\%$ \\
     & & $4\times 6400$   &-& $58.4\%$ \\
     & & $16\times 6400$  &-& ${60.2\%}$ \\
    
    \midrule
    \multicolumn{5}{l}{\textit{Tree Search Methods}} \\
    \midrule
    {PACT   \citep{han2021proof}} & {837M} & $1\times16\times512$ & $23.9\%$ & $24.6\%$ \\
    &  & $8\times16\times512$ & $29.3\%$ & $29.2\%$ \\
     ReProver   \citep{yang2024leandojo} & 229M & -& - & $26.5\%$ \\
     Llemma   \citep{azerbayev2023llemma} & 7B & $1\times32\times100$ & $26.2\%$ & $26.2\%$ \\
     Llemma   \citep{azerbayev2023llemma} & 34B & $1\times32\times100$ & $27.9\%$ & $25.8\%$ \\

    {Curriculum Learning   \citep{polu2022formal}} & {837M} & $1\times8\times512$ & $33.6\%$ & $29.6\%$ \\
     &  & $8\times8\times512$ & $41.2\%$ & $34.5\%$ \\
     &  & $64\times8\times512$ & $47.3\%$ & $36.6\%$ \\
     {HTPS   \citep{lample2022hypertree}} & {600M} & cumulative & $58.6\%$ & - \\
     &  & $64\times5000$ & - & $41.0\%$ \\
     Lean-STaR ~\citep{lin2024leanstarlearninginterleavethinking}& 7B & $64\times1\times50$ & - & $46.3\%$ \\

     InternLM2-Math~\citep{ying2024internlmmathopenmathlarge} & 7B &  $1\times32\times100$ & $29.9\%$ & $30.3\%$ \\
     InternLM2-Math-Plus & 7B & $1\times32\times100$ &- & $43.4\%$  \\
    DeepSeek-Prover-V1.5-RL & {7B} &  $1\times3200$ &-& $55.0\%$ \\
     & &  $4\times6400 $ &-& $59.6\%$ \\
     & & $16\times6400$ &-& ${62.7\%}$ \\
     & & $32\times6400$ &-& ${63.5\%}$ \\
     InternLM2-StepProver \citep{wu2024leangithubcompilinggithublean}    & {7B} &   $1\times32\times100$ (beam) & $59.8\%$ & $48.8\%$  \\
     &&  $64\times32\times100$ & ${63.9\%}$ & ${54.5\%}$\\    
     \midrule
     {\thiswork -BF   } & {7B} &  $1\times32\times600$ & $55.4\%$ & $47.3\%$  \\
                                                 && $4\times32\times600$ & ${61.3\%}$  & ${52.6\%}$\\   
                                                 && $16\times32\times600$ & ${63.7\%}$ & ${57.3\%}$\\  
                                                 && $64\times32\times600$ & ${64.6\%}$ & ${59.2\%}$\\ 
                                                 && $256\times32\times600$ & ${65.1\%}$ & ${59.4\%}$\\ 
     \midrule
          {\thiswork -CG   } & {7B} &  $1\times32\times600$ & $49.4\%$ & $43.0\%$  \\
                                                 && $4\times32\times600$ & ${55.9\%}$  & ${56.1\%}$\\   
                                                 && $16\times32\times600$ & ${64.5\%}$ & ${61.7\%}$\\  
                                                 && $64\times32\times600$ & ${67.7\%}$ & ${64.3\%}$\\ 
                                                 && $256\times32\times600$ & ${68.4\%}$ & ${65.6\%}$\\ 
     \midrule
     {\thiswork -BF+CG   } & {7B} &  $2\times32\times600$ & $56.0\%$ & $50.7\%$  \\
                                                 && $4\times32\times600$ & ${61.4\%}$  & ${58.5\%}$\\   
                                                 && $16\times32\times600$ & ${65.8\%}$ & ${62.5\%}$\\  
                                                 && $64\times32\times600$ & ${68.0\%}$ & ${63.8\%}$\\ 
                                                 && $256\times32\times600$ & $\textbf{69.6\%}$ & $\textbf{65.9\%}$\\ 
    \bottomrule
\end{tabular}
\end{center}
\end{table*}

\paragraph{miniF2F}
We first analyze the performance on the miniF2F benchmark~\cite{zheng2021minif2f}. The original benchmark was released in Lean 3 and was later ported to an earlier version of Lean 4. We use the Lean 4 version of miniF2F, as released by the LeanDojo project~\cite{yang2024leandojo}, with our adaptations to Lean 4.7.0 and corrections of several formalization mistakes.

The best-first-search approach employs an evaluation setting similar to that of InternLM2-StepProver, where the model selects states to expand based on the average log-likelihood of the tactics leading to those states. In contrast, the critic-guided (CG) search method involves the prover model selecting states to expand based on \thiswork-critic that grades each state. The search budget for both methods can be universally described as \( P \times S \times K \), where \( P \) represents the number of passes, \( S \) the number of states, and \( K \) the maximum number of state expansions, or search iterations. In our context, we set \( S = 32 \) and \( K = 600 \), with the temperature fixed at \( T = 0.7 \). In a BF+CG scenario, the computation budgets are equally distributed. 
The test results are presented in Tab.~\ref{tab:main_results}. 


\paragraph{ProofNet}
The evaluation setting is mainly the same as miniF2F. Tab.~\ref{tab:proofnet_results} presents the performance of various models on the ProofNet dataset. The BF+CG strategy equally distributes search budgets on two methods. As we are not taking the validation set of ProofNet for expert iteration, only the pass rate on the whole dataset is reported. \thiswork\ achieved a pass@256 of 27.0\% overall.

\paragraph{Critic-Guided Search Improves Theorem Proving.} From Tab.~\ref{tab:main_results}, it is observed that critic-guided search performs significantly better than the best-first method using the same prover model. The critic-guided method achieves an accuracy rate of 68.4\% on miniF2F-Valid and 65.6\% on miniF2F-Test, which is only slightly lower than the combined methods and substantially higher than the BF method. These experimental results indicate that the critic model enhances the inherent proving ability of the prover model, enabling it to discover deeper proofs. Analysis of the length distribution of the searched solutions (shown in Fig.~\ref{fig:proof_depth}) further validates this observation, as CG methods consistently discover deeper proofs than BF methods, successfully constructing proofs with more than 9 tactics.


\paragraph{Critic-Guided Search Explores a Distinct Proof Space.}
Another observation is that the proof distribution discovered by critic-guided (CG) search differs significantly from that of the best-first (BF) approach. The synergistic effect of combining the two methods highlights this distinction. For instance, a hybrid model achieves 69.6\% accuracy on miniF2F-Valid and 65.9\% on miniF2F-Test. On the ProofNet benchmark, this synergy is even more pronounced: while BF and CG individually solve 22.3\% and 23.9\% of problems, their combination solves 27.0\%. These results strongly indicate that critic-guided search navigates the proof space differently, identifying novel solutions that a scaled-up BF search alone would not discover. However, with low sampling budgets, CG may overlook simpler proofs that are trivial for BF. Therefore, a hybrid search was adopted to leverage the strengths of both methods in our experiment.
\begin{table*}[t]
\setlength{\tabcolsep}{0.06in}
\begin{center}
\small
\caption{\centering
Pass rates on the ProofNet \citep{azerbayev2023proofnetautoformalizingformallyproving} dataset. }
\begin{tabular}{lcccc}
\toprule
\multirow{2}{*}{Method} & \multirow{2}{*}{Pass} & \multicolumn{3}{c}{ProofNet} \\
 & & valid & test & all \\
\toprule
ReProver \citep{yang2024leandojo} & - & - & - & $13.8\%$ \\

\multirow{2}{*}{Deepseek-Prover-V1.5-RL} & $1\times 3200$ & $22.0\%$ & $21.5\%$ & $21.8\%$ \\
 & $4\times 6400$ & $25.4\%$ & $25.3\%$ & $25.3\%$ \\
InternLM2-StepProver & $1\times 32\times 100$ & - & - & $18.1\%$ \\
\midrule
\multirow{3}{*}{\thiswork-BF }
& $2\times 32\times 600$ & - & - & $18.8\%$ \\
& $32\times32\times 600$ & - & - & $21.1\%$ \\
& $128\times32\times 600$ & - & - & \textbf{$22.3\%$} \\
\multirow{3}{*}{\thiswork-CG }
& $2\times 32\times 600$ & - & - & $17.4\%$ \\
& $32\times32\times 600$ & - & - & $21.9\%$ \\
& $128\times32\times 600$ & - & - & \textbf{$23.9\%$} \\
\multirow{3}{*}{\thiswork-BF+CG }
& $4\times 32\times 600$ & - & - & $18.8\%$ \\
& $64\times32\times 600$ & - & - & $23.6\%$ \\
& $256\times32\times 600$ & - & - & \textbf{$27.0\%$} \\
\bottomrule
\end{tabular}

\label{tab:proofnet_results} 
\end{center}
\end{table*}



\subsection{Results of Expert Iteration}
As Tab.~\ref{tab:leanwkbk_results} shows, a total of 17.0\% of the Lean-workbook-plus problems are proved or disproved, making a noticeable improvement since the release of Lean-Workbook.
These proved and disproved statements and their corresponding tactics and states have been released.
We also revealed more facts about the expert iteration process, especially the efficacy of our CG method.

\begin{table*}[h]
\setlength{\tabcolsep}{0.06in}
\begin{center}
\small
\caption{\centering
Results on the Lean-workbook-plus \citep{ying2024leanworkbooklargescalelean} dataset.
}
\begin{tabular}{lcccc}
\toprule
\multirow{2}{*}{Method} & \multirow{2}{*}{Pass} & \multicolumn{3}{c}{Lean-workbook-plus} \\
 && Proved & Disproved & Total \\
\toprule
 InternLM2-StepProver & cumulative & 7,909 (9.5\%) & - & 7,909 (9.5\%) \\
 \thiswork & cumulative & 10,880 (13.1\%) & 3,195 (3.9\%) & \textbf{14,075 (17.0\%)}\\
\bottomrule
\label{tab:leanwkbk_results}
\end{tabular}
\end{center}
\end{table*}

\paragraph{Scaling CPU/GPU Computation in Expert Iteration.} The search process involves a collaboration of CPU and GPU resources. Given the fixed amount of active GPUs and CPUs, the GPU time consumed is proportional to the total CPU time (in our case, approximately 1:11). In summary, approximately 21,364 CPU days are consumed throughout the entire expert iteration process. However, these search budgets are not uniformly distributed across all formalized problems. Easier problems are more likely to be solved in the early rounds of iteration, thereby ceasing to consume search resources in later rounds. The search consumption of each problem is a key indicator of the distribution of problem difficulty and can provide valuable insights for further scaling. In our case, we selected the CPU time consumed per successful proof as an estimate of resource consumption expectations. Consider the set of problems \( S = \{s_i\} \). For each problem \( s_i \), let \( P_{s_i} \) denote the set of all attempts, and \( T_{s_i,j} \) represent the time spent on the \( j \)-th attempt of problem \( s_i \). Define the indicator function \( \text{valid}(s_i,j) \), which equals 1 if the \( j \)-th attempt on problem \( s_i \) results in a valid solution, and 0 otherwise. The CPU time consumed per successful proof, \( C_{s_i} \), is given by:
$C_{s_i} = \frac{\sum_{j} T_{s_i,j}}{\sum_{j} \text{valid}(s_i,j)}$.

Table~\ref{tab:leanwkbk_cpu_results} presents a detailed analysis of computational resource consumption. A key observation is that the majority of CPU resources are expended on problems that are challenging to prove or disprove. Notably, only about 1.5\% of CPU resources are used to solve 17.0\% of the problems, while the remaining 98.5\% of resources yield no successful outcomes.

\begin{figure}[t]
	\centering
        \includegraphics[width=0.9\linewidth]{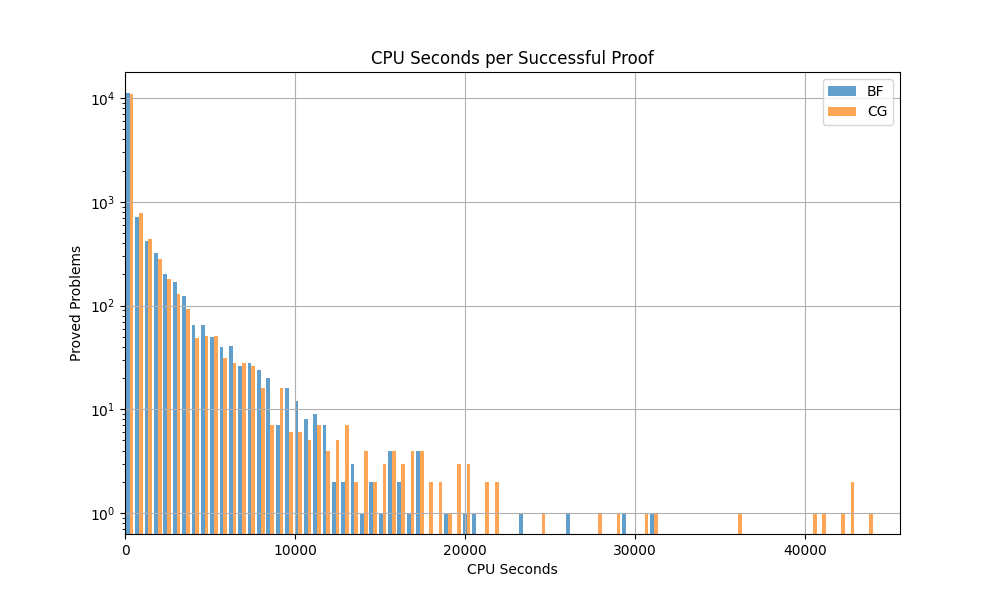}
        
		\caption{The distribution of CPU search time per successful proof, which is defined as the total CPU resources spent on searching the specific problem divided by the number of independent successful search trials. Only problems solved are represented in the figure. 83\% of the Lean-workbook problems that remain unproven are not depicted. BF and CG are analyzed independently.}
		\label{fig:time-distribution}
	\vspace{-0.15in}
\end{figure}

Fig~\ref{fig:time-distribution} provides a more granular analysis of the distribution of CPU search time for each problem. A number of observations can be made from the graph. First, the graph reveals a peak in the near-zero region, suggesting the presence of numerous trivial problems in the auto-formalized dataset. Second, a log-linear trend is evident for problems with CPU search times between approximately 0 and 10,000 seconds. Assuming an equal distribution of problem difficulties, searching for longer proofs becomes exponentially harder due to the search space explosion problem, which explains the occurrence of this trend. Unfortunately, the best-first approach is observed to degrade to a brute-force approach—discovering fewer proofs with larger search budgets (>10,000 s)—while the critic-guided approach performs better, discovering more solutions beyond the log-linear trend.

\begin{table}[h]
    \setlength{\tabcolsep}{0.06in}
    \begin{center}
    \small
    \caption{\centering
    CPU time spent on the Lean-workbook\citep{ying2024leanworkbooklargescalelean} dataset.
    }
    \begin{tabular}{lcc}
    
    \toprule
    {Problem State} & Number & {Total CPU days}\\
    \toprule
     Proved/Disproved & 14,075 (17.0\%) & 331 (1.5\%)\\
     Remain unproven & 68,200 (83.0\%) & 21,033 (98.5\%) \\
    \bottomrule
    \end{tabular}
    
    \label{tab:leanwkbk_cpu_results} 
    \end{center}
\end{table}


\begin{figure}[t]
		\includegraphics[width=1\linewidth]{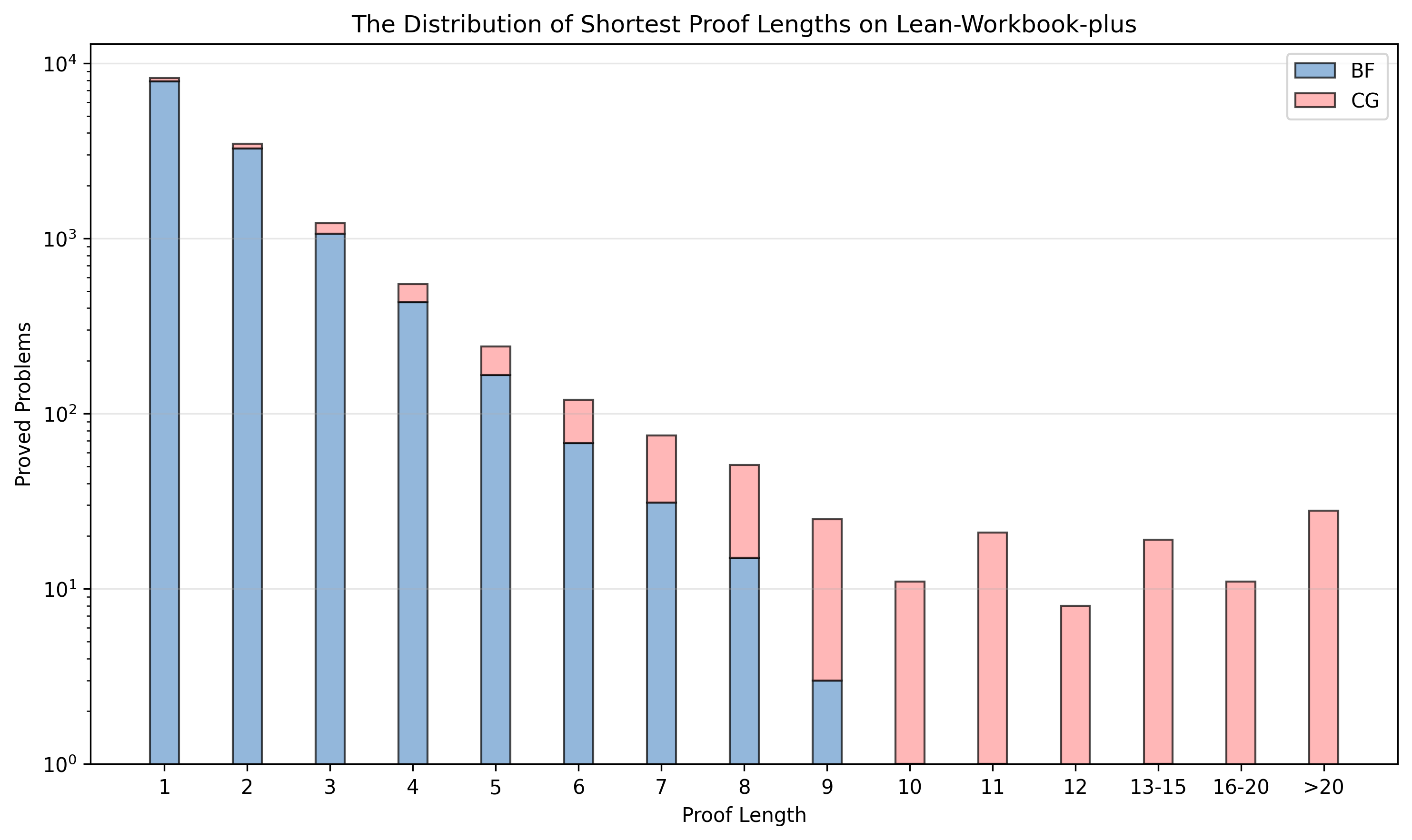}
		\caption{The distribution of shortest proof length.}
		\label{fig:len-distribution}
\end{figure}

\paragraph{Scaling Proof Lengths in Expert Iteration.} As analyzed in the previous section, best-first-search (BF) methods suffer from an explosion of the search space, rendering it inferior under large sampling budgets. In contrast, critic-guided (CG) methods are capable of finding longer solutions. The average length of solutions found by the BF method is 1.66, whereas the same indicator is 4.44 for the CG method, demonstrating a significant improvement in deeper reasoning abilities. The distribution of proofs during our expert iteration is shown in Fig.~\ref{fig:len-distribution}. To avoid the impact of redundancies, only the shortest proof for each problem is considered here. A nearly log-linear trend is observed in the distribution of proof lengths found by best-first search method, while the critic-guided proof method can find much longer proofs. However, 83\% of the problems in the Lean-workbook remain unproven, which can have longer proof lengths and require much more time to find them based on this estimation. We hope that our proposed method can help facilitate the scaling of formal proof searching beyond the "log-linear" boundaries.

\section{Related Work}
\paragraph{Automatic Theorem Proving.}
Automated theorem proving does not have a unified approach.
A mainstream paradigm is training a language model on tuples of (proof state, next tactic), followed by a tree search to find proofs \citep{polu2020generative, polu2022formal, lample2022hypertree, yang2024leandojo, lin2024leanstarlearninginterleavethinking,wu2024leangithubcompilinggithublean}.
Another line of work is training to auto-regressively generate a whole proof based on a theorem statement from the model itself \citep{xin2024deepseekproveradvancingtheoremproving, xin2024deepseekproverv15harnessingproofassistant} or translating from human informal proofs \citep{jiang2022draft,wu2022autoformalization,wang2024theoremllamatransforminggeneralpurposellms}.
Regardless of the learning paradigm used, most methods rely on expert iteration \citep{expertit} to improve the model and require the prover model itself to determine the search direction. In this work, we train our prover model to search under the guidance of an external critic model. The models are trained with a state-tactic paradigm via expert iteration on a large-scale LEAN dataset.

\paragraph{Process-supervised Reasoning in Informal Mathematics.}
Recently, numerous studies have shown that allowing an external language model to evaluate the quality of reasoning steps can boost performance on tasks such as informal math~\cite{lightman2023letsverifystepstep, zheng2025processbenchidentifyingprocesserrors, uesato2022solvingmathwordproblems}. Although this approach has proven effective, it often requires a significant amount of human-annotated data or extensive computation~\cite{luo2024improvemathematicalreasoninglanguage, wang2024mathshepherdverifyreinforcellms}. Furthermore, such methods are restricted to informal language reasoning tasks. We propose a new scheme for training a tactic-level critic model that leverages the correctness signal from the formal system.

\section{Conclusion}
In this paper, we presented critic-guided search, a novel approach that significantly enhances the theorem proving capabilities of prover language models in formal mathematics by using an external critic model to rerank intermediate Lean states. Our method begins with training initial critic models from the trajectories of prior prover models to predict preference information between Lean states. We further improved the prover and the critic in tandem using expert iteration, fine-tuning the models on correct proof trajectories that the prover model samples and verifies using the Lean solver. 
 Our contributions include the introduction of the theorem
 proving dataset Lean-Workbook-Plus, demonstrating that critic-guided search can further improve search depth, and achieving
 new results on the miniF2F-test benchmark, increasing the pass rate from 54.5\% to 65.9\%.

These advancements not only improve the performance of automated theorem proving, but also offer concrete guidance for future developments in automated mathematical reasoning.

\section*{Limitations}
This work is mainly focused on contest-level math problems and pays less attention to other automated theorem-proving scenarios. Besides, we currently do not have a stable metric to measure critic models, which makes iteration of critic models difficult.

\section*{Impact Statement}
This paper presents work whose goal is to advance the field of 
Machine Learning. There are many potential societal consequences 
of our work, none of which we feel must be specifically highlighted here.

\nocite{langley00}

\bibliography{example_paper}
\bibliographystyle{icml2025}

\newpage
\appendix
\onecolumn
\def\lstlanguagefiles{lstlean.tex}

\section{Training details}
\paragraph{Prover Model} Our prover model is built upon InternLM-math-plus-7B~\cite{ying2024internlmmathopenmathlarge}. We used the same training setting when we performed the expert iteration process: We used a global batch size of 512 and a learning rate of $2\times10^{-5}$. We fine-tuned for 2 epochs to obtain the SFT model. For the learning rate, we used a warm-up in the first 3\% steps, followed by a cosine schedule decaying to zero. The entire expert iteration process generated 2.19 billion tokens of data, with the final iteration taking approximately 14 hours on 32 A800 GPUs.

\paragraph{Critic Model} We initialize the critic model from InternLM2-Chat-1\_8b-sft\footnote{\url{https://huggingface.co/internlm/internlm2-chat-1_8b-sft}}\citep{internlm2} and fine-tune it for one epoch.
We create preference pairs among miniF2F-valid \citep{zheng2021minif2f}, Mathlib \citep{mathlib}, and Lean-Workbook-Plus \citep{ying2024leanworkbooklargescalelean} using best-first-search. The final-round data includes 454K pairs where we have removed duplicate pairs and reduced the number of pairs containing \texttt{no\_goals} to 10\% of their original count. 
We train critic models with 8 A800 GPUs.
We evaluate our critic model using preference pairs generated on the miniF2F-test with 6510 pairs.
We use the accuracy metric defined as the proportion of correctly predicted positive and negative pairs. The model achieved an accuracy of 78.0\%, demonstrating its preliminary ability to distinguish between positive and negative pairs in the proof tree.

\section{Case studies}
Here we list interesting cases proved by \thiswork\ from different datasets.

\begin{tcolorbox}[
colback=white!10!white,
colframe=purple!75!purple,
title=Case: Lean-workbook
]
\textcolor{blue}{Natural Language problem:}  For natural numbers \( m \) and \( n \), if \((mn + m + n) \mod 6 = 4\), then \( 12 \mid mn \).
\\

\begin{lstlisting}[language=LEAN]
theorem lean_workbook_plus_74374 (m n : ℕ) : (m * n + m + n) % 6 = 4 → 12 | m * n   :=  by
simp [Nat.add_mod, Nat.mul_mod, Nat.mod_mod]
rw [← Nat.mod_add_div m 6, ← Nat.mod_add_div n 6]
have h₁ : m % 6 < 6 := Nat.mod_lt _ (by norm_num)
have h₂ : n % 6 < 6 := Nat.mod_lt _ (by norm_num)
interval_cases m % 6 <;> interval_cases n % 6 <;> simp_all (config := {decide := true})
all_goals ring_nf; simp [Nat.dvd_iff_mod_eq_zero, Nat.mul_mod, Nat.add_mod, Nat.mod_mod]
\end{lstlisting}
\label{tab:mathd_case_formal}
\end{tcolorbox}
\thiswork\ successfully addresses the problem by imposing constraints on the range of variables and then solving it directly using enumeration techniques. This example illustrates the distinction between formal and informal reasoning styles.

\begin{tcolorbox}[
colback=white!10!white,
colframe=purple!75!purple,
title=Case: miniF2F: mathd\_algebra\_31
]
\textcolor{blue}{Natural Language problem:} If $ \sqrt{x+\!\sqrt{x+\!\sqrt{x+\!\sqrt{x+\cdots}}}}=9$, find $x$. Show that it is 72.
\\

\begin{lstlisting}[language=LEAN]
theorem mathd_algebra_31 (x : NNReal) (u : ℕ → NNReal) (h₀ : ∀ n, u (n + 1) = NNReal.sqrt (x + u n))
    (h₁ : Filter.Tendsto u Filter.atTop (mathscrN 9)) : 9 = NNReal.sqrt (x + 9) := by 
    have h₂ := h₁.const_add x
    have h₃ : Filter.Tendsto (fun k => NNReal.sqrt (x + u k)) Filter.atTop (mathscrN (NNReal.sqrt (x + 9))) :=
      NNReal.continuous_sqrt.continuousAt.tendsto.comp h₂
    have h₄ : (fun k : ℕ => NNReal.sqrt (x + u k)) = fun k : ℕ => u (k + 1) := by
      ext k
      rw [h₀]
    have h₅ : Filter.Tendsto (fun k : ℕ => u (k + 1)) Filter.atTop (mathscrN (NNReal.sqrt (x + 9))) :=
      h₄ ▸ h₃
    have h₆ : Filter.Tendsto (fun k => u (k + 1)) Filter.atTop (mathscrN 9) := h₁.comp (Filter.tendsto_add_atTop_nat 1)
    exact tendsto_nhds_unique h₆ h₅
\end{lstlisting}
\label{tab:mathd_case_formal}
\end{tcolorbox}
This case demonstrates how \thiswork~solves a problem whose formalized version is significantly harder than the informal one. The informal solution of this problem is not rigorous, which jumps from the equation $ \sqrt{x+\!\sqrt{x+\!\sqrt{x+\!\sqrt{x+\cdots}}}}=9$ to $\sqrt{x+9}=9$, involving a substitution that is intuitive but risky. The formalized version of this problem uses series and limitations to redefine the problem. This is a case where formal reasoning detaches from informal reasoning. In such cases, it is hard to say that the involvement of informal CoT has any benefit to the problem. \thiswork\ provides a solid proof of the problem without the augmentation of informal information, successfully solving the problem.

\begin{tcolorbox}[
colback=white!10!white,
colframe=purple!75!purple,
title=Case: ProofNet: Munkers\_31\_2
]
\textcolor{blue}{Natural Language problem:} Show that if $X$ is normal, every pair of disjoint closed sets has neighborhoods whose closures are disjoint.
\\

\begin{lstlisting}[language=LEAN]
theorem exercise_Munkers_31_2 {X : Type*}
[TopologicalSpace X] [NormalSpace X] {A B : Set X}
(hA : IsClosed A) (hB : IsClosed B) (hAB : Disjoint A B) :
∃ (U V : Set X), IsOpen U ∧ IsOpen V ∧ A ⊆ U ∧ B ⊆ V ∧ closure U ∩ closure V = ∅ := by
obtain ⟨U₀, V₀, hU₀, hV₀, hA₀, hB₀, hAB₀⟩ := normal_separation hA hB hAB
obtain ⟨U, hU₁, hU₂, hU₃⟩ := normal_exists_closure_subset hA hU₀ hA₀
obtain ⟨V, hV₁, hV₂, hV₃⟩ := normal_exists_closure_subset hB hV₀ hB₀
refine ⟨U, V, hU₁, hV₁, hU₂, hV₂,?_⟩
exact (hAB₀.mono hU₃ hV₃).eq_bot
\end{lstlisting}
\label{tab:mathd_case_formal}
\end{tcolorbox}
\thiswork\ has an improved capability of solving undergraduate problems, even though it was not fine-tuned on such data distributions. \thiswork\ utilizes premises from the Mathlib to construct valid closure subsets, effectively demonstrating its mathematical reasoning ability.

\end{document}